\documentclass[runningheads]{llncs}
\usepackage{graphicx}
\usepackage{amsmath}
\usepackage{amsfonts}
\usepackage{amssymb}
\usepackage{multirow}
\usepackage{caption}
\usepackage{subcaption}
\usepackage{epstopdf}
\usepackage{placeins}
\usepackage{fancyhdr}
\usepackage{csquotes}
\usepackage{adjustbox}
\usepackage{float}
\usepackage{algorithm}
\usepackage{algorithmic}
\usepackage{arydshln}
\usepackage{verbatim}
\usepackage{newfloat}
\usepackage{listings}
\usepackage{xcolor}
\usepackage{hyperref}
\hypersetup{
    colorlinks=true,
    linkcolor=cyan
    }
\newcommand{\repeatthanks}{\textsuperscript{\thefootnote}}
\begin{document}
\title{A Unified Framework for Slot based Response Generation in a Multimodal Dialogue System}
%
%
\author{Mauajama Firdaus\thanks{denotes equal contribution by both authors} \inst{1} \and
Avinash Madasu\repeatthanks \inst{2} \and
Asif Ekbal\inst{3}}
\authorrunning{Firdaus et al.}
%
\institute{Department of Computing Science, University of Alberta, Alberta, Canada
\email{mauzama.03@gmail.com}\\
\and
Cognitive Computing Research, Intel Labs, Santa Clara, California, USA
\email{avinashmadasu17@gmail.com}\\
 \and
Computer Science and Engineering, Indian Institute of Technology Patna, Patna, Bihar, India\\
\email{asif@iitp.ac.in}}
\maketitle              
\begin{abstract}
Natural Language Understanding (NLU) and Natural Language Generation (NLG) are the two critical components of every conversational system that handles the task of understanding the user by capturing the necessary information in the form of slots and generating an appropriate response in accordance with the extracted information. 
Recently, dialogue systems integrated with complementary information such as images, audio, or video have gained immense popularity. In this work, we propose an end-to-end framework with the capability to extract necessary slot values from the utterance and generate a coherent response, thereby assisting the user to achieve their desired goals in a multimodal dialogue system having both textual and visual information. The task of extracting the necessary information is dependent not only on the text but also on the visual cues present in the dialogue. Similarly, for the generation, the previous dialog context comprising multimodal information is significant for providing coherent and informative responses. We employ a multimodal hierarchical encoder using pre-trained DialoGPT and also exploit the knowledge base (Kb) to provide a stronger context for both the tasks. Finally, we design a slot attention mechanism to focus on the necessary information in a given utterance. Lastly, a decoder generates the corresponding response for the given dialogue context and the extracted slot values. Experimental results on the Multimodal Dialogue Dataset (MMD) show that the proposed framework outperforms the baselines approaches in both the tasks.  The code is available at \url{https://github.com/avinashsai/slot-gpt}.

\end{abstract}
\section{Introduction}\label{sec1}
Advancement in Artificial Intelligence (AI) has opened up new frontiers in conversational agents. Human-machine interaction is an essential application of AI helping humans in their day-to-day lives. Progress in AI has led to the creation of personal assistants like Apple's Siri, Amazon's Alexa, and Microsoft's Cortana which assist humans in their everyday work. The machines' capability to comprehend and complete the user's goals has empowered researchers to build advanced dialogue systems.
Dialogue systems with the ability to help users solve critical tasks such as selecting the appropriate restaurants, and hotels, or booking movie tickets have gained immense popularity in the field of artificial intelligence (AI). Through a well-designed conversation system as an efficient personal assistant, users can easily achieve everyday tasks through natural language interactions. 

With the growth in AI, the latest progress in deep learning has encouraged many neural conversational systems ~\cite{serban2016building,wen2017network,li2017end}. A typical goal-oriented dialogue system comprises several key modules such as: (i) Natural language understanding (NLU) component that helps in identifying the domain, intent and extract slot information from the user utterance \cite{williams2019neural,xia2018zero,chen2019self,niu2019novel}; (ii). a dialogue state tracker (DST) that predicts the current dialogue state \cite{zhong2018global,rastogi2017scalable}; (iii). a dialogue policy that regulates the next system action given the current state \cite{wang2020modelling,peng2018deep}; (iv). a natural language generator (NLG) module that outputs a response given the semantic frame \cite{shang2015neural,serban2017hierarchical,tian2019learning,madasu2022unified}. These modules occur in a pipeline in every robust dialogue system. Therefore, it is slightly time-consuming and computationally expensive. 

With the progress in AI, the integration of information from different modalities, such as text, image, audio, and video, has been known to provide complete information for building effective end-to-end dialogue systems \cite{saha2018towards,le2019end,madasu2022learning,madasu2023improving}
by bringing the different areas of computer vision (CV) and natural language processing (NLP) together. 
Hence, a multimodal dialogue system bridges the gap between vision and language, ensuring interdisciplinary research. 

Multimodal conversational systems provide completeness to the existing dialogue systems by providing necessary information that lacks in unimodal systems as the visual (in the case of images and videos) and audio information help build robust systems. In \cite{saha2018towards} the authors proposed a multimodal dialogue dataset having textual and image information for the fashion domain. From the dataset, it is clear that image information is necessary for selecting the right clothes and accessories for different individuals.

\begin{figure}[!h]
    \centering
    \begin{adjustbox}{max width=\linewidth}
    \includegraphics{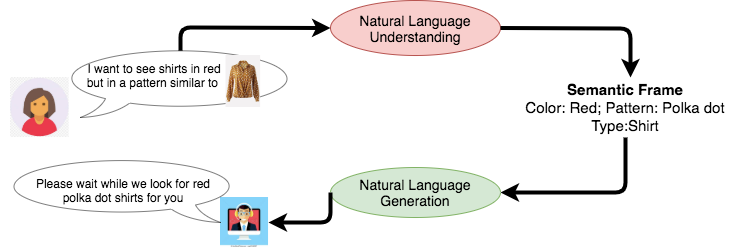}
    \end{adjustbox}
    \caption{NLU and NLG modules in a Dialogue System}
    \label{example}
\end{figure}

\subsection{Motivation and Contribution}
As demonstrated in Figure \ref{example}, the primary goal of every NLU component is to extract necessary information in the form of slots from the user utterance, while the ultimate goal of the NLG module is to respond to the user based on the extracted semantic information. Both these tasks are complementary; hence information extracted from the NLU is significant for generating the correct response by the NLG unit. Instead of performing these tasks separately in a pipeline manner, recently, researchers have focused on performing these tasks simultaneously to improve the performance of both tasks \cite{su2019dual,dong2019unified}. 

We take a step forward in our current work by proposing an end-to-end system that can concurrently extract the necessary slot information from the user utterance and provide the corresponding system response in a multimodal dialogue setting. This is more challenging as the slot information is not entirely dependent on the textual utterance but also on the visual information. Hence, for a better generation of responses, extraction of the correct semantic information from the current dialogue context is crucial.

Slots are crucial as it provides the key semantic information for a better understanding of the user utterance. To provide informative responses to the user it is important to capture the semantic information in the form of slots. Based on the slot values, the response generation module can provide responses that are informative and engaging. The proposed end-to-end framework first captures the slot information and then uses this slot information captured from both text and image as input for the generation module.

The key contributions of our current work are three-fold: 
\begin{itemize}
    \item We propose the task of simultaneously performing two critical components of every conversational system, i.e. NLU and NLG, in a multimodal dialogue system employing information from both text and images.
    \item We design a slot attention-based hierarchical generation system using pre-trained DialoGPT.
    \item Our proposed system achieves the best performance compared to the existing and baseline approaches in both tasks. 
\end{itemize}

The rest of the paper is structured as follows. In Section 2, we present a brief review of the existing literature. We provide the details of the baseline and the proposed methodology in Sections 3 and 4 respectively. In Section 5, we provide the details of the dataset used and its statistics followed by implementation details and evaluation metrics. Experimental results are presented in Section 6 along with a detailed analysis, including error analysis. Finally, in Section 7 we conclude with future directions of research. 

\section{Related Work}
In any dialogue framework, Natural Language Generation (NLG) is a classic problem. With the fast growth of Artificial Intelligence (AI), there has been a trend in recent times to develop multimodal dialogue systems by combining text with images, audio and video modalities. A brief description of some of the works carried out in unimodal chatbots, accompanied by multimodal dialogue systems for both the tasks of slot filling and response generation, is provided below.

\subsection{Slot Filling}
Several deep learning architectures have also been employed for extracting essential information in the form of slots from a given utterance. The authors in \cite{deoras2013deep} investigated deep belief networks (DBN) for slot filling on the ATIS dataset. In \cite{mesnil2013investigation}, the authors investigated Elman and Jordan-type RNNs for slot filling. In \cite{mesnil2015using}, several hybrid variants of RNN were proposed due to the stronger ability of RNNs to capture dependencies compared to traditional models, such as Conditional Random Field (CRF). In \cite{yao2013recurrent} lexical, syntactic and word-class features were used as input to an RNN for the SLU task of slot filling. 

The authors in \cite{yao2014recurrent} used the transition features to improve RNNs and the sequence level criteria for optimisation of CRF to capture the dependencies of the output label explicitly. The authors in \cite{yao2014spoken} used deep LSTMs along with regression models to obtain the output-label dependency for slot filling. The usage of kernel deep convex networks (K-DCN) was investigated in \cite{deng2012use} for slot filling. In \cite{zhu2017encoder}, a focus mechanism for an encoder-decoder framework was proposed for slot filling on the ATIS dataset. The authors in \cite{zhao2018improving} introduced a generative network based on the sequence-to-sequence model along with a pointer network for slot filling. In \cite{shin2018slot}, an attention-based encoder-decoder framework has been employed for slot filling.

In \cite{qiu2018recurrent}, a pre-trained language model was employed in an RNN framework for the slot-filling task. Attention-based RNN framework was proposed in \cite{wu2018attention} along with pre-trained word embeddings for identifying the slots on ATIS and MEDIA datasets. On the ATIS dataset in \cite{lan2018semi}, an adversarial multi-task model combining a bi-directional language model with a slot tagging model was used for identifying the slots in a given user utterance. 

The adversarial framework was used in \cite{liu2017multi} for learning common representation across multiple domains for slot-filling tasks. In \cite{zhu2018concept}, the authors proposed the concept of transfer learning for the task of slot filling as it is an essential task of language understanding.
Authors in \cite{williams2019neural} encoded lexicon information as features for use in a Long-short term memory neural network for slot-filling tasks. With advancements in AI, multimodality has been incorporated into conversational systems to make them more robust and complete. Recently, authors in \cite{zhang2019neural} used an adaptive attention mechanism to extract the necessary slot values in a multimodal dialogue system.

\subsection{Response Generation}
\subsubsection{Unimodal Dialogue System:}
The effectiveness of deep learning clearly shows significant improvements in dialogue generation. Deep neural models are very effective in modelling the dialogues, as seen in \cite{vinyals2015neural,shang2015neural}. In \cite{sordoni2015neural}, a context-sensitive neural language approach was presented where, given the textual conversational background, the model chooses the most likely answer. To capture the context of the previous queries by the users, the authors in \cite{sordoni2015hierarchical} proposed a hierarchical framework capable of preserving past information. Sequence-to-sequence (seq2seq) neural models often generate incomplete and boring responses, such as \enquote{I don't know}, \enquote{Okay}, \enquote{Yes}, \enquote{No}, etc. Hence, bringing diversity in responses is an extremely challenging and interesting research problem for every conversational agent. 

Similarly, to preserve the dependencies among the utterances, a hierarchical encoder-decoder framework was investigated in \cite{serban2016building,serban2017hierarchical}. The authors in \cite{xu2019end} extended the hierarchical encoder-decoder framework by adding a latent variable for understanding the intentions of the conversations in a task-oriented dialog system. Lately, memory networks \cite{madotto2018mem2seq} have been intensely investigated for capturing the contextual information in dialogues for the generation of responses infusing pointer networks. 

Hierarchical pointer networks \cite{raghuhierarchical} has also been employed for response generation in task-oriented dialogues. The authors in \cite{wu2019global} incorporated a global encoder and a local decoder to share external knowledge in a task-oriented dialogue setup. The ability to infuse knowledge in responses was achieved by using a Bag-of-sequence memory unit \cite{raghu2018disentangling} for generating coherent responses in goal-oriented dialogue systems. The authors in \cite{reddy2018multi} proposed a multi-level memory framework for task-oriented dialogues. A memory-augmented framework with the ability to extract meaningful information during training for better response generation has been explored in \cite{tian2019learning}. With the release of MultiWoz \cite{budzianowski2018multiwoz}, a task-oriented dialogue dataset, several works have focused on multi-domain dialogue generation.

The authors in \cite{budzianowski2019hello} used a pre-trained language model for dialogue generation. A hierarchical graph framework employing the dialogue acts of the utterances was investigated for dialogue generation in \cite{chen2019semantically}. The meta-learning approach \cite{mi2019meta,qian2019domain} has been applied to different datasets to increase the domain adaptability for generating the responses. 
To increase the ability to memorize the dialogue context, the authors in \cite{wu2019learning} used a memory-to-sequence framework and the pointer generator for response generation. A multi-task framework to enhance the performance of natural language generation was investigated in \cite{zhu2019multi}. 

In \cite{chen2019working}, working memory was employed for dialogue generation. The working memory interacts with two long-term memories that capture the dialogue history and the knowledge base tuples for the informative response generation. Recently, a heterogeneous memory network \cite{lin2019task} has been explored for response generation having the capability to simultaneously use the dialogue context, user utterance, and the knowledge base for response generation. Dynamic fusion technique has been employed in \cite{qin2020dynamic} to share the features across different domains for a better generation.

\subsubsection{Multimodal Dialogue System:}
Research in the dialogue system has recently shifted towards incorporating different sources of information, such as images, audio, video, and text in order to make a robust system. The research reported in \cite{das2017visual,mostafazadeh2017image,de2017guesswhat,gan2019multi,firdaus2020emosen} has been useful in narrowing the gap between vision and language. In \cite{mostafazadeh2017image}, an Image Grounded Conversations (IGC) task was proposed, where conversations are natural and focused upon a shared image. Similarly, the authors in \cite{das2017visual} introduced the task of visual dialogue, which requires an AI agent to hold a meaningful dialogue with humans in natural, conversational language about the visual content. 

Recently, video and textual modalities were investigated with the release of the DSTC7 dataset in \cite{le2019multimodal} that used a multimodal transformer network to encode videos and incorporate information from the different modalities. Similarly in \cite{le2019end,alamri2018audio,lin2019entropy}, the DSTC7 dataset has been used for generation by incorporating audio and visual features. The release of the Multimodal dialogue (MMD) dataset \cite{saha2018towards}, having conversations on the fashion domain in both text and images, has facilitated response generation in multimodal setup. 

Several works on the MMD dataset reported in \cite{agarwal2018knowledge,agarwal2018improving,liao2018knowledge} used the hierarchical encoder-decoder model to generate responses by capturing information from text, images, and the knowledge base. Recently, \cite{chauhan2019ordinal} proposed attribute-aware and position-aware attention for generating textual responses. The authors in \cite{cui2019user} used a  hierarchical attention mechanism for generating responses on the MMD dataset. In \cite{firdaus2020more}, the authors proposed a stochastic method for generating diverse responses in a multimodal dialogue setup. Multi-domain multi-modal aspect controlled response generation task was introduced in \cite{firdaus2020multidm}. 

Lately, the authors have focused on jointly addressing NLU and NLG tasks in a unimodal framework \cite{tseng2020generative,su2019dual,dong2019unified} for improving the performance of both tasks. Author's in \cite{tseng2020generative} proposed a generative model which couples NLU and NLG through a shared latent variable. Similarly, in \cite{su2019dual} a new learning framework was designed for language understanding and generation on top of dual supervised learning, providing a way to exploit the duality. 

Our current work differs from the existing NLU and NLG works as we intend to build a comprehensive framework that extracts the necessary slot information and generates the appropriate response adhering to the elicited slot information in a multimodal framework. The task becomes more complex as visual cues in the form of images are also crucial for providing the complete context for both the tasks along with the textual information.

\section{Methodology}
In this section, we discuss the problem statement followed by the baseline and the proposed methodology.

\subsection{Problem Definition}
In this paper, we address the task of extracting the slot values from the user utterance and generating informative and relevant textual responses according to the conversational history in a multimodal dialogue setting.  
The dialogues consist of textual utterances along with multiple images. 
More precisely, given a user utterance $U_p = u_{p,1}, u_{p,2}, ... , u_{p,j}$, a set of images $I_p = img_{p,1}, img_{p,2}, ... , img_{p,j'}$, with the dialogue history $H_p = (U_1,I_1), (U_2,I_2), ..., (U_{p-1},I_{p-1})$, we focus on extracting the slot information from $U_p$ and $I_p$ and simultaneously generate interesting, informative, context-aware response $Y_p = (y_{p,1}, y_{p,2}, \ldots, y_{p,k}$) instead of template like generic and monotonous responses, such as \textit{I don't know, Yes, No, Similar to...}, etc. This will enhance human-machine conversations by keeping the users engaged in the conversation. Here, $p$ is the $p^{th}$ turn of a given dialogue, while $j$ is the number of words in a given textual utterance and $j'$ is the number of images in a given utterance. Note that in every turn, the number of images $j'$ $\leq$ 5, so in the case of only text, vectors of zeros are considered in place of image representation.

\begin{figure*}[htbp]
    \centering
    \begin{adjustbox}{max width=\linewidth}
    \includegraphics{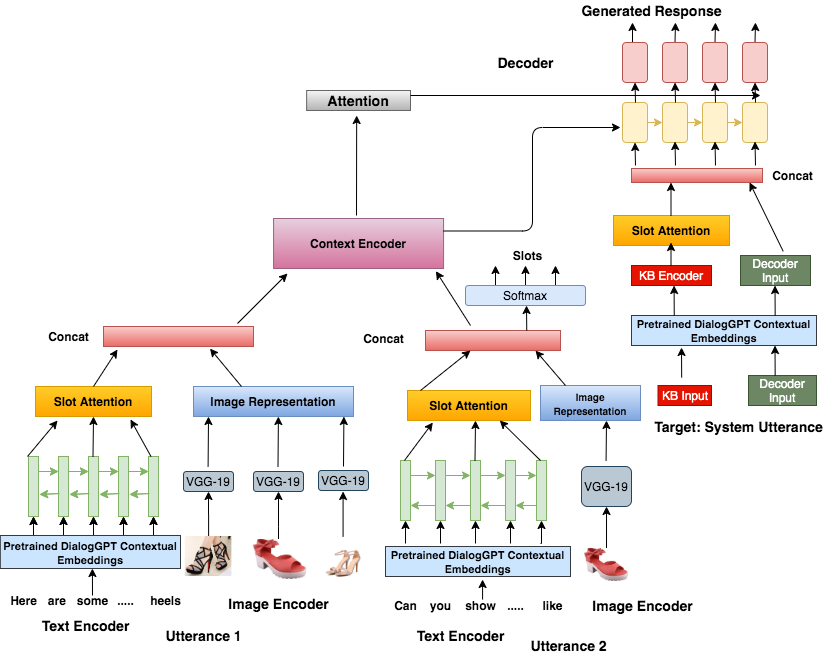}
    \end{adjustbox}
    \caption{Architecture of the proposed model}
    \label{arch}
\end{figure*}

\subsection{Multimodal Hierarchical Encoder Decoder:} \label{sec:mhred}
We construct a generative model for response generation, an extension of the recently introduced Hierarchical Encoder-Decoder (HRED) architecture \cite{serban2017hierarchical,serban2016building}. As opposed to a standard sequence-to-sequence model \cite{sutskever2014sequence}, the dialogue context among the utterances is captured by adding utterance-level RNN (Recurrent Neural Network) over the word-level RNN, increasing the efficacy of the encoder to capture the hierarchy in dialogue. The multimodal HRED (MHRED) is built upon the HRED to include text and image information in a single framework. The critical components of MHRED are the utterance encoder, image encoder, context encoder, and decoder.

\subsubsection{Utterance Encoder:}
Given an utterance $U_p$, we use bidirectional Gated Recurrent Units (BiGRU) \cite{cho2014learning} to encode each word $n_{p,i}$, where $i \in (1,2,3,.....k)$ having $d$-dimensional embedding vectors into the hidden representation $h_{U,p,i}$ as follows:

\begin{equation}
\begin{split}
    \overrightarrow {h_{U,p,i}}= GRU_{U,f} ({n_{p,i}},\overrightarrow{h_{U,p,i-1}})\\
    \overleftarrow {h_{U,p,i}}= GRU_{U,b} ({n_{p,i}},\overleftarrow{h_{U,p,i-1}})\\
    h_{U,p,k}^{txt} = [\overrightarrow {h_{U,p,i}}, \overleftarrow {h_{U,p,i}}]
    \end{split}
\end{equation}

here $\overrightarrow {h_{U,p,i}}$ represent the utterance representation in the forward direction while $\overleftarrow {h_{U,p,i}}$ represents in the backward direction. The overall representation of the utterance is given by $h_{U,p,k}^{txt}$.

\subsubsection{Image Encoder:}
A pre-trained VGG-16 \cite{simonyan2014very} having a 16-layer deep convolutional neural network (CNN) trained on more than one million images present in the ImageNet dataset is used for encoding the images. It can classify images into 1000 object categories, such as dresses, shoes, animals, keyboards, mouse, etc. As a result, the network can learn rich features from a wide range of images. Here, it is also used to extract the \enquote{local} image representation for all the images in the dialogue turns and concatenates them together. The concatenated image vector is passed through the linear layer to form the global image context representation as given below:

\begin{equation}
\begin{split}
     T_{p,i} = VGG(I_{p,i}) 
\\
         T_p = Concat(T_{p,1} , T_{p,2}, \ldots, T_{p,j’})
\\
     h_{I,p}^{img} = ReLU(W_IT_p + b_I)
     \end{split}
\end{equation}

where $W_I$ and $b_I$ are the trainable weight matrix and biases. In every turn, the number of images $i \leq$ 5, so in the case of only text, vectors of zeros are considered in place of image representation. 

\subsubsection{Context Encoder:}
The final hidden representations from both image and text encoders are concatenated for each turn and are given as input to the context-level GRU. A hierarchical encoder is built to model the conversational history on top of the image and text encoder. The decoder GRU is initialized by the final hidden state of the context encoder.

\begin{equation}
    h_{w,p}^{ctx} = GRU_w([h_{U,p,k}^{txt};h_{I,p}^{img}], h_{w,p-1})
\end{equation}

where $h_{w,p}^{ctx}$ is the final hidden representation of the context for a given turn.

\subsubsection{Decoder:}
In the decoding section, we build another GRU for generating the words sequentially based on the hidden state of the context GRU and the previously decoded words. We use input feeding decoding and the attention \cite{luong2015effective} mechanism for enhancing the performance of the model. Using the decoder state $h_{d,t}^{dec}$ as the query vector, the attention layer is applied to the hidden state of the context encoder. The context vector and the decoder state are concatenated and used to calculate a final distribution of probability over the output tokens.

\begin{equation}
\begin{split}
    h_{q,t}^{dec} = GRU_d(y_{p,t-1}, h_{q,{t-1}});\\
    \alpha_{t,m} = softmax({{h_{w,p}^{ctx}}^T}{W_f}{h_{q,t}})\\
    c_t = \sum_{m=1}^{k} {\alpha_{t,m}} {h_{w,p}^{ctx}}, \\
    \tilde{h}_t = tanh(W_{\tilde{h}}{[{h_{q,t}};c_t]});\\
    P(y_t/y_{<t}) = softmax({W_S}{\tilde{h}_t})
    \end{split}
\end{equation}
where, $W_f$, $W_S$ and $W_{\tilde{h}}$ are trainable weight matrices.

\subsection{Proposed Approach:}
To further improve the MHRED model's performance, we propose to apply slot attention to the utterances. The goal is to focus on the slot values in a user utterance crucial in generating an appropriate system response. If the model fails to attend to vital slot information, then it generates an inappropriate system response. For example, for the given user utterance \enquote{\textit{Will \textbf{neck-tie} having \textbf{25 cm} size be paired well with any of these?}}, the slot values \textbf{neck-tie} and \textbf{25 cm} are crucial to understanding user utterance. 

Furthermore, in a Multimodal Dialogue system, the user often refers to the images generated in the previous system responses. The model should account for this subtle but very essential information. For example, in the following user utterance, \enquote{\textit{Show me more in style as in the \textbf{4th} image}}, the model must understand that the user is referring to the \textbf{4th} image. Any failure in doing so will generate an inapt system response. Hence, in our proposed approach, we employ mechanisms to improve the performance of slot identification. The architecture of the proposed model is shown in Figure \ref{arch}.

\subsubsection{Slot Attention:} \label{sec:slot}
We employ self-attention on the output from Utterance Encoder $h_{U,p,k}^{txt}$ as in the final Equation 1. We refer to Slot Attention as SA in the rest of the paper.
Let  $h_{U,p,k}^{txt}$ be the Key (K), Query (Q) and Value (V).
\begin{equation}
    SA(K, Q, V) = softmax(\frac{QK^{\rm T}}{\sqrt{d_{k}}})V.
\end{equation}
where $d_{k}$ is the hidden dimension size of $h_{U,p,k}^{txt}$.
The output from Slot Attention is concatenated with the Image Encoder's output. The concatenated output is sent as an input to the Context Encoder.

\subsubsection{Knowledge Base (KB):} \label{sec:kb}
The knowledge base encoder used in our framework is the same as \cite{agarwal2018knowledge}. The knowledge base of the MMD dataset contains information about contextual queries and celebrities endorsing various products and brands. Hence, to provide this additional information to our proposed model, we employ self-attention on KB input to achieve more focused information as follows:
\begin{equation}
\begin{split}
    h_s^{query} = n_p^{query} (h_{n-1}^{query}, k_{l,t})
\\
    h_f^{ent} = n_p^{ent} (h_{n-1}^{ent}, d_{l,t})
\\
     h_{net}^{kb} = [h_s^{query},h_f^{ent}]
         \end{split}
\end{equation}
Let  $h_{net}^{kb}$ be the Key (K), Query (Q) and Value (V).
\begin{equation}
    SA(K, Q, V) = softmax(\frac{QK^{\rm T}}{\sqrt{d_{k}}})V.
\end{equation}
where $d_{k}$ is the hidden dimension size of $h_{net}^{kb}$.
We use the attended KB output and the decoder input as the combined input at each time step of the decoder.

As the knowledge base (KB) input remains intact for a particular dialogue context, we concatenate the KB input with the decoder input in a similar manner as \cite{agarwal2018knowledge}. 

\subsubsection{Pretraining DialoGPT (P-GPT):}
In a dialogue system, understanding contextual information is crucial to performance enhancement. Pre-trained language models have achieved state-of-the-art results on several Natural Language Understanding (NLU) tasks \cite{radford2019language}. Furthermore, pre-training dialogue systems significantly improved the performance of generation \cite{ju2019all}. Therefore, we pre-train Multi-modal Dialog (MMD) dataset using DialoGPT \cite{zhang2019dialogpt}. The input to the DialoGPT is a combination of the previous system response, the current user utterance, and the current system response. This helps the model to learn long-range contextual information effectively. We use DialoGPT-small for a context size of 2. The pre-trained contextual embeddings are passed as input to the Text Encoder and KB Encoder.

\subsubsection{Slot Prediction:}
For slot prediction, we take the scores obtained by applying the softmax layer on the output of the dot product between Query (Q) and Key (K) in the self-attention for a given user utterance to find the distribution on the slot values for a given user utterance. 

\subsubsection{Training and Inference:}
The generation model is trained using teacher-forced cross entropy \cite{williams1989learning} at every decoding step to minimize the negative log likelihood on the model distribution. We define $\hat{y} =  {\hat{y}_1, \hat{y}_2, \hat{y}_3, \ldots,  \hat{y}_m }$ as the ground truth of the given input sequence.
\begin{equation}
    J(\theta) = -\sum_{s=1}^n logp ({\hat{y}_t} | y_1 ,y_2, \ldots, y_{m-1} ) 
\end{equation}
 here $y_1$, $y_2$ $\ldots$ and $y_{m-1}$ is the generated utterance. 

\section{Comparison Methods}
In this section, we provide a comparison with the other existing techniques comprising both state-of-the-art methods and other baselines. 
\subsection{State-of-the-Art Models:}

\textbf{Seq2Seq:}
It is an encoder-decoder framework with attention which is a standard baseline in Machine Translation, Generation \cite{sutskever2014sequence}. The input to the encoder is dialogue history and the decoder output is the next round-generated dialogue. 

\textbf{HRED:}
It is the first hierarchical encoder-decoder architecture proposed for text-based dialogue systems and is a standard baseline for Unimodal system \cite{serban2015hierarchical}. It also follows a similar input-output format as the Seq2Seq model.

\textbf{MHRED:}
Multimodal hierarchical encoder-decoder is the first model proposed for Multimodal dialogue systems. Along with text, image is also served as input \cite{saha2018towards}. The input to the encoder is a concatenated input of image and text features and the decoder output is the generated dialogue.

\textbf{UMD:}
A user-guided attention model is proposed to consider hierarchical product taxonomy and users’ attention to products. It is based upon MHRED architecture \cite{cui2019user}. The attention model focuses on user preferences for the products and generates responses based on them.

\textbf{OAM:}
In this paper, a novel position and attribute-aware attention mechanism are proposed to learn the enhanced image representation conditioned on
the user utterance. The proposed model can generate appropriate
responses while preserving the position and attribute information \cite{chauhan2019ordinal}.

\textbf{MAGIC:}
Multimodal diAloG system with adaptIve deCoders (MAGIC) first judges response through understanding user intention. It then applies an adaptive decoder for generating apposite responses \cite{nie2019multimodal}.

\textbf{MATE:}
It is based on the standard transformer architecture. In the encoding stage, the transformer encoder is used to encode information from multimodal input. Generation is a two-stage process based on the transformer decoder. In the first stage, the focus is more on the encoded information. In the second stage, responses are refined by incorporating domain knowledge into the output of the first stage \cite{he2020multimodal}.

\textbf{LXMERT:} 
In LXMERT \cite{tan2019lxmert}, the authors build a large-scale Transformer model that consists of three encoders: an object relationship encoder, a language encoder, and a cross-modality encoder. We concatenate all the images and feed them as input to the visual encoder (i.e., object-relationship encoder) and the language encoder is used for utterance representation. Finally, the cross-modality encoder is used to capture the final utterance representation for both text and images. 

\subsection{Baseline Models:} 
To show the effectiveness of the proposed components, we implement the models without these components in the architecture.\\
\textbf{Unimodal Baselines:}
In order to prove that multimodal architectures perform better, we compare these with unimodal architectures. In unimodal architectures, only text served as the input to the models.\\
\textbf{Without Kb:}
We experiment with the models without using the Knowledge base as inputs at the decoder. It is used as a baseline to compare with the models using the Knowledge base as input.\\
\textbf{Without Slot Attention:}
In these models, Slot Attention (SA) is not applied. It is used as a comparison to demonstrate the efficiency of the Slot Attention component.\\
\textbf{Without Pretrained Dialog-GPT (P-GPT) representations:}
To prove our hypothesis that Pretrained Dialog-GPT improves performance, we perform experiments without Pretrained Dialog-GPT representations as input. It is to show the performance differentiation between the models using P-GPT and without using P-GPT.

\section{Dataset and Experiments}
\label{sec:length}
In this section, we provide the details of the datasets used for experiments, implementation details, evaluation metrics and the results obtained.   
\subsection{Dataset Description:}
Our research is based on the Multi-modal Dialog (MMD) dataset \cite{saha2018towards} \footnote{https://amritasaha1812.github.io/MMD/download/} consisting of 150k chat sessions between the customer and sales representative. During the sequence of customer-agent interactions, domain-specific information in the fashion domain was collected. The dialogues easily integrate text and image knowledge into a conversation that brings together different modalities to create a sophisticated dialogue system. 

The dataset presents new challenges for multi-modal, goal-oriented dialogue systems having complex user sentences. The detailed information of the MMD dataset is presented in Table \ref{data}. The authors \cite{saha2018towards}, for experimentation \enquote{unroll} the different images 
to incorporate only one image for a single utterance. 
Though computationally learns, the method eventually lacks the goal of capturing multi-modality over the context of multiple images and text.
Therefore, in our study, we use a different version of the dataset as outlined in \cite{agarwal2018improving,agarwal2018knowledge} to capture a large number of images as the concatenated context vector for each turn of a dialogue. 
The motivation behind this is that multiple images are required to provide the correct responses to the users. 
\begin{table}[h]
\centering
\begin{adjustbox}{max width=\linewidth}
\begin{tabular}{|c|ccc|}
\hline
\textbf{Dataset Statistics} & \textbf{Train} & \textbf{Valid} & \textbf{Test} \\ \hline
\textit{Number of dialogues} & 105,439 & 22,595 & 22,595 \\
\textit{Avg. turns per Dialogue} & 40 & 40 & 40 \\
\textit{\begin{tabular}[c]{@{}c@{}}No. of Utterances with\\ Text Response\end{tabular}} & 1.54M & 331K & 330K \\
\textit{\begin{tabular}[c]{@{}c@{}}Avg. words in Text \\ Response\end{tabular}} & 14 & 14 & 14 \\ 
\textit{\begin{tabular}[c]{@{}c@{}}No. of Utterances with\\  Image Response\end{tabular}} & 904K & 194K & 193K \\ \hline
\end{tabular}
\end{adjustbox}
\caption{Dataset statistics of Multi-modal Dialogue (MMD) dataset}\label{data}
\end{table}


\subsection{Implementation Details}
\subsubsection{DialoGPT Pretraining}
We used the DialoGPT-small model for pre-training. Previous system response, current user utterances and current system response are concatenated together into a single sentence separated by a special token. The goal is to capture long range contextual information. Adam is used as the optimizer with a learning rate of 5e-5 and a batch size of 16. The pretraining is performed for 3 epochs. The maximum value of the gradient norm is 1.0.
The model's weights are initialized with the already pre-trained weights in DialoGPT paper \cite{zhang2019dialogpt}. The loss function used is the same as the one used in the DialoGPT paper.

\subsubsection{Model Training}
All the implementations are done using the PyTorch\footnote{https://pytorch.org/} framework. The input embedding dimensions are 512 for randomly initialized word embeddings and 768 for pre-trained contextual embeddings. The hidden size for all the layers is 512. A dropout \cite{srivastava2014dropout} of 0.3 is applied on the Slot Attention for all T-HRED models and 0.5 for all M-HRED models. All the models are trained for 15 epochs with a batch size of 256. AdamW \cite{loshchilov2017decoupled} is used as the optimizer with a learning rate of 0.0001 for all the models. For image representation, a 4096-dimensional FC6 layer from VGG-19 network \cite{simonyan2014very} is used, which is trained on ImageNet.

\subsection{Automatic Evaluation Metric:}
To evaluate the model at the relevance and grammatical level, we report the results using standard metrics like Rouge-L \cite{rouge2004} and BLEU-1,2,3 and 4 \cite{Papineni2002BleuAM}. For comparison with the existing approaches, we report NIST metric \cite{doddington2002automatic} in a similar manner as \cite{he2020multimodal}. To evaluate the slot value extraction performance, we use the traditional metrics such as F1 score and Accuracy similarly as \cite{williams2019neural,zhang2019neural}. For accuracy and F1 scores, the percentage values in the range >=70\% and <80\%- represent fair results; while $>=$80\% and $<$90\% represents good models; and $>=$90\% represents very good/excellent models.

\subsection{Human Evaluation Metrics}
We recruit six annotators (in a similar manner as \cite{shang2015neural,tian2019learning}) from a third-party company, having high-level language skills. We sampled 500 responses per model for evaluation with the utterance and the conversational history provided for a generation.
First, we evaluate the quality of the response on two conventional criteria: \textit{Fluency} and \textit{Relevance}. We also compute slot consistency for our proposed task which determines whether the generated response is consistent with the predicted slot information. These are rated on a five-scale, where 1, 3, and 5 indicate unacceptable, moderate, and excellent performance, respectively, while 2 and 4 are used for unsure. 
We compute Fleiss' kappa \cite{fleiss1971measuring} to measure inter-rater consistency. The Fleiss' kappa for Fluency and Relevance is 0.53 and 0.49, indicating moderate agreement. For Slot Consistency, we obtain 0.65 as the kappa score indicating substantial agreement. 

\begin{table}[h]
\centering
\begin{adjustbox}{max width=\linewidth}
\begin{tabular}{c|c|cc}
\hline
\multicolumn{2}{c|}{\textbf{Model Description}} & \textbf{Accuracy} & \textbf{F1 Score} \\ \hline
\multirow{2}{*}{\textbf{\begin{tabular}[c]{@{}c@{}}Unimodal Baselines\end{tabular}}} 
 & \textit{P-GPT HRED + SA} & 57.8 & 57.4 \\
 & \textit{P-GPT HRED + Kb + SA} & 59.4 & 58.6 \\ \hline
\multirow{1}{*}{\textbf{\begin{tabular}[c]{@{}c@{}}Multimodal Baselines\end{tabular}}} 
 & \textit{P-GPT MHRED + SA} & 60.5 & 59.2 \\ \hline
\multirow{3}{*}{\textbf{\begin{tabular}[c]{@{}c@{}}Proposed Approach\end{tabular}}} & \textit{\begin{tabular}[c]{@{}c@{}}\textit{P-GPT MHRED + SA} + \textit{Kb}\end{tabular}} & \textbf{62.4} & \textbf{61.3} \\
& \textit{\begin{tabular}[c]{@{}c@{}}\textit{P-GPT MTrans + SA} + \textit{Kb}\end{tabular}} & \textbf{65.1} & \textbf{63.8} \\
& \textit{\begin{tabular}[c]{@{}c@{}}\textit{P-GPT Mul-Trans + SA} + \textit{Kb}\end{tabular}} & \textbf{66.3} & \textbf{64.7} \\\hline
\end{tabular}
\end{adjustbox}
\caption{Evaluation results on Slot information. Here, P-GPT is the Pre-trained DialoGPT}  
\label{slot_res}
\end{table}

\begin{figure*}[htbp]
\centering
        \begin{subfigure}[b]{0.5\textwidth}
                \centering
                \includegraphics[width=\linewidth]{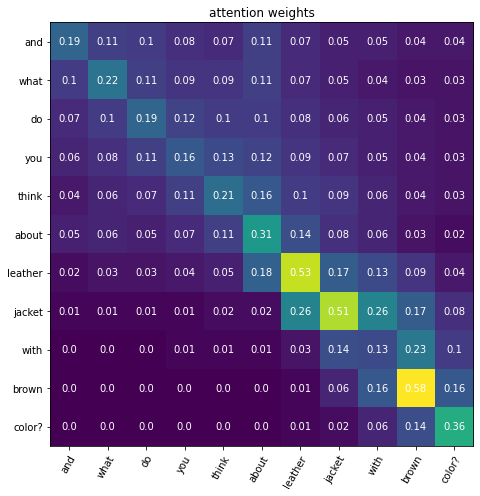} 
                \caption{Example 1}
                \label{eg1}
        \end{subfigure}\hfill
        \begin{subfigure}[b]{0.5\textwidth}
                \centering
                \includegraphics[width=\linewidth]{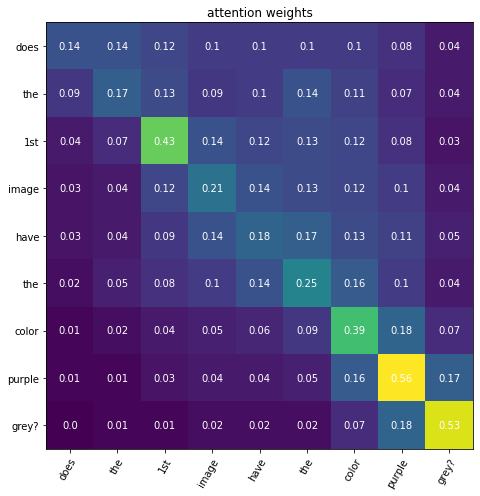} 
                \caption{Example 2}
                \label{eg2}
        \end{subfigure}\hfill
  \begin{subfigure}[b]{0.5\textwidth}
                \centering
                \includegraphics[width=\linewidth]{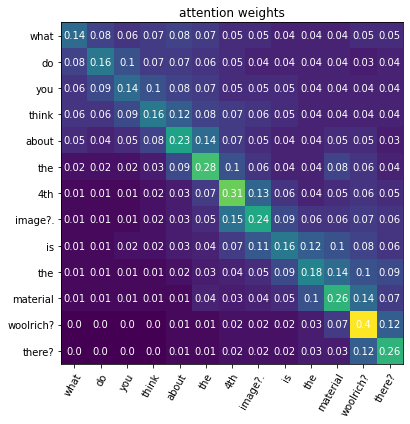}  
                \caption{Example 3}
                \label{eg3}
        \end{subfigure}\hfill 
         \begin{subfigure}[b]{0.5\textwidth}
                \centering
                \includegraphics[width=\linewidth]{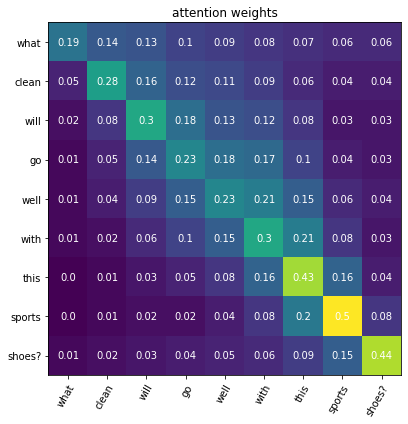}  
                \caption{Example 4}
                \label{eg4}
        \end{subfigure}\hfill   
\caption{Attention visualization }\label{atten_slot} \end{figure*}
\section{Result and Analysis}
This section presents the experimental results for both the tasks and the necessary analysis of the baseline models generated responses and the proposed methodology.

\subsection{Slot Prediction Results}
The results of the slot values are provided in Table \ref{slot_res}. This shows that the proposed approach outperforms all the existing baseline models (both unimodal and multimodal), and these improvements are statistically significant. 
With the addition of external knowledge, the slot performance improves significantly, thereby assisting the framework in capturing the correct values from the given user utterance. Besides, the pre-trained DialoGPT embeddings help provide a stronger context for the task with a gain of approximately 3\% from the baseline \textit{MHRED + Kb} model. This justifies that both the pre-trained embeddings and external knowledge, along with slot attention, are crucial in providing the slot information for correct extraction and finally assist in a generation. 

In Figure \ref{atten_slot}, we provide a few attention visualizations of the user utterances to show that the slot attention designed to extract the correct values is efficient in capturing the correct information. The slot attention in Example 1 has been able to focus on the colour \textit{brown} and material \textit{leather}, providing the right slot values. Similarly, in Example 2 the proposed framework has correctly attended the colour information \textit{purple grey} and position of the image $1^{st}$ in the given user utterance. 

By utilizing transformers as an encoder we see a boost in performance with an improvement of 2\% in F1 score and around 3\% in accuracy for the slot prediction task. The obvious reason behind the improvement is the ability of the transformer network to better capture the contextual information of a given utterance in comparison to the MHRED model which uses GRU as the basic cell for encoding the utterance. 

The proposed approach achieves a performance gain of more than 6\% and 11\% in accuracy with respect to the unimodal baselines having the knowledge information and without knowledge respectively. Similarly, by adding the multimodal information using the transformer networks there is an increase of 3\% compared to \textit{P-GPT MHRED + SA + Kb} framework. This proves that the transformer approach captures better semantic knowledge as opposed to the encoder-decoder method.
\begin{table}[h]
\centering
\begin{adjustbox}{max width=\linewidth}
\begin{tabular}{c|c|c|c|ccccc}
\hline
\multicolumn{2}{c|}{\textbf{Model Description}} & \textbf{SA} & \textbf{P-GPT} & \textbf{BLEU-1} & \textbf{BLEU-2} & \textbf{BLEU-3} & \multicolumn{1}{l}{\textbf{BLEU-4}} & \multicolumn{1}{l}{\textbf{Rouge-L}} \\ \hline
\multirow{6}{*}{\textbf{\begin{tabular}[c]{@{}c@{}}Unimodal\\ Baselines\end{tabular}}} & \multirow{3}{*}{\textit{HRED}} & - & - & 0.624 & 0.535 & 0.475 & 0.425 & 0.666 \\
 &  & $\surd$ & - & 0.635 & 0.544 & 0.483 & 0.433 & 0.668 \\  
 &  & $\surd$ & $\surd$ & 0.638 & 0.547 & 0.486 & 0.436 & 0.671 \\ \cline{2-9} 
 & \multirow{3}{*}{\textit{HRED + Kb}} & - & - & 0.646 & 0.560 & 0.503 & 0.456 & 0.685 \\
 &  & $\surd$ & - & 0.657 & 0.565 & 0.505 & 0.460 & 0.682 \\  
 &  & $\surd$ & $\surd$ & 0.659 & 0.571 & 0.507 & 0.460 & 0.683 \\ \hline
\multirow{3}{*}{\textbf{\begin{tabular}[c]{@{}c@{}}Multimodal\\ Baselines\end{tabular}}} & \multirow{2}{*}{\textit{MHRED}} & - & - & 0.630 & 0.541 & 0.478 & 0.43 & 0.669 \\
 &  & $\surd$ & - & 0.636 & 0.545 & 0.484 & 0.434 & 0.668 \\ 
 &  & $\surd$ & $\surd$ & 0.638 & 0.547 & 0.487 & 0.437 & 0.672 \\ \hline 
\multirow{9}{*}{\textbf{\begin{tabular}[c]{@{}c@{}}Proposed\\ Approaches\end{tabular}}} & \multirow{3}{*}{\textit{MHRED + Kb}} & - & - & 0.649 & 0.563 & 0.503 & 0.455 & 0.685 \\
&  & $\surd$ &- & 0.659 & 0.571 & 0.512 & 0.463 & 0.688 \\
 &  & $\surd$ & $\surd$ & \textbf{0.662} & \textbf{0.574} & \textbf{0.514} & \textbf{0.465} & \textbf{0.690} \\ \cline{2-9} 
 & \multirow{3}{*}{\textit{MTrans + Kb}} & - & - & 0.665 & 0.569 & 0.515 & 0.462 & 0.693 \\
&  & $\surd$ & - & 0.673 & 0.579 & 0.527 & 0.470 & 0.711 \\
 &  & $\surd$ & $\surd$ & \textbf{0.684} & \textbf{0.587} & \textbf{0.535} & \textbf{0.479} & \textbf{0.724} \\ \cline{2-9}
 & \multirow{3}{*}{\textit{Mul-Trans + Kb}} & - & - & 0.671 & 0.575 & 0.524 & 0.473 & 0.704 \\
&  & $\surd$ & - & 0.679 & 0.582 & 0.531 & 0.481 & 0.713 \\
 &  & $\surd$ & $\surd$ & \textbf{0.690} & \textbf{0.591} & \textbf{0.544} & \textbf{0.493} & \textbf{0.733} \\ \hline
\end{tabular}
\end{adjustbox}
\caption{Results using automatic evaluation metrics. Here, P-GPT is the pre-trained DialoGPT;  HRED, MHRED, HRED + Kb, and MHRED + Kb framework without SA and P-GPT is similar to \cite{agarwal2018improving,agarwal2018knowledge} respectively. }  
\label{auto_res}
\end{table}

\subsection{Response Generation Results}
\subsubsection{Results on Automatic Evaluation:}
In Table \ref{auto_res}\footnote {we perform statistical significance t-test \cite{welch1947generalization}, and it is conducted at 5\% (0.05) significance level}, we present the results of automatic evaluation. As already stated, we report BLEU 1,2,3 and 4 as BLEU measures the n-grams overlap between the generated response and the gold response that would help measure if the extracted slot values improve the performance of generation. From the  results shown in the table, it is evident that our proposed framework performs significantly better in comparison to the baseline models. As the primary objective was to ensure that slot extraction helps in a better generation, the results justify that by capturing the correct values by the slot attention mechanism, we achieve better performance in the case of all the metrics. There is a definite improvement of 1\% in the case of baseline and the proposed framework by incorporating slot attention. This proves that slot knowledge makes the generated response more informative by adding the correct slot values in the responses.

Multimodality in the form of images plays a crucial role in building robust systems. In our framework, visual information in the form of images has been incorporated, and from the table, it is obvious that there is a slight improvement in the performance as opposed to the unimodal frameworks having only textual information. Finally, we employ pre-trained DialoGPT embeddings to enhance the performance of the overall generation process. From the table, it is visible that by utilizing the pre-trained embeddings, there is a gain in performance in the proposed approach and the baselines. In particular, there is approximately 1 point improvement in the final model with DialoGPT embeddings compared to the framework without any pre-trained embedding. This ensures that pre-training is beneficial in capturing better context, thereby providing stronger dialogue information to generate informative and coherent responses. The multimodality information provided more complementary information that is not presented by the textual modality thereby improving the overall performance.

By employing transformer along with a knowledge base without having the slot information and pre-trained embeddings, we see that it performs better than the \textit{MHRED + Kb} model in terms of Rouge-L and BLEU scores. By adding slot information and pre-trained embedding, the performance of the model improves significantly. 
In \textit{Mul-Trans + Kb} framework, we use multimodal transformers in the sense that for utterance representation we use transformers while the image representation achieved from VGG-19 is fed as input to a transformer network in a similar manner as \cite{akbari2021vatt}. Here, the utterance information from the transformer along with the output of the image representation from the transformers is concatenated to get the context of the entire utterance representation having both textual and visual knowledge. 
The representation achieved from both transformers is then used for generating the response. 

As evident from Table \ref{auto_res}, the model with transformer representations for both textual and visual representation achieves performance improvement over the \textit{MTrans + Kb} framework that uses a transformer network only for utterance representation. We also compare our framework with the LXMERT \cite{tan2019lxmert}, model and we see that it performs better compared to all the baselines still our proposed network outperforms LXMERT. This is due to the fact that LXMERT captures object-oriented features accompanied by captions for a single image. But in our case, we don't have captions and also the images are multiple in number compared to the LXMERT framework. Also, certain utterances do not have visual information in most of the dialogues.

Evidently, the performance of the \textit{Mul-Trans + Kb} model is significantly better as opposed to the RNN networks due to the capability and efficacy of the transformers in capturing better-contextualized representations using multi-head attention and feed-forward networks. The Rouge-L score and BLEU scores are the highest compared to all the baselines for the proposed \textit{Mul-Trans + Kb} model.
\begin{table}[h]
\centering
\begin{adjustbox}{max width=\linewidth}
\begin{tabular}{c|c|cccc|c}
\hline
\multicolumn{2}{c|}{\multirow{2}{*}{\textbf{Model}}} & \multicolumn{4}{c|}{\textbf{BLEU}} & \multirow{2}{*}{\textbf{NIST}} \\ \cline{3-6}
\multicolumn{2}{c|}{} & \textbf{1} & \textbf{2} & \textbf{3} & \textbf{4} &  \\ \hline
\multirow{2}{*}{\textbf{\begin{tabular}[c]{@{}c@{}}Unimodal\\ Baselines\end{tabular}}} & \textit{\textbf{Seq2Seq} \cite{sutskever2014sequence}} & 35.39 & 28.15 & 23.81 & 20.65 & 3.3261 \\
 & \textit{\textbf{HRED} \cite{serban2015hierarchical}} & 35.44 & 26.09 & 20.81 & 17.27 & 3.1007 \\ \hline
\multirow{6}{*}{\textbf{\begin{tabular}[c]{@{}c@{}}Multimodal\\ Baselines\end{tabular}}} & \textit{\textbf{MHRED} \cite{saha2018towards}} & 32.60 & 25.14 & 23.21 & 20.52 & 3.0901 \\
 & \textit{\textbf{UMD} \cite{cui2019user}} & 44.97 & 35.06 & 29.22 & 25.03 & 3.9831 \\
 & \textit{\textbf{OAM} \cite{chauhan2019ordinal}} & 48.30 & 38.24 & 32.03 & 27.42 & 4.3236 \\
 & \textit{\textbf{MAGIC} \cite{nie2019multimodal}} & 50.71 & 39.57 & 33.15 & 28.57 & 4.2135 \\
 & \textit{\textbf{MATE} \cite{he2020multimodal}} & 56.55 & 47.89 & 42.48 & 38.06 & 6.0604 \\
& \textit{\textbf{LXMERT} \cite{tan2019lxmert}} & 64.32 & 51.33 & 45.33 & 42.76 & 7.3855 \\
\hline
\multirow{6}{*}{\textbf{\begin{tabular}[c]{@{}c@{}}Proposed\\ Approach\end{tabular}}} & \textit{\textbf{\begin{tabular}[c]{@{}c@{}}P-GPT + MHRED + SA + \\ (Joint Training)\end{tabular}}} & \textbf{66.20} & \textbf{57.40} & \textbf{51.40} & \textbf{46.50} & \textbf{6.3164} \\ 
& \textit{\textbf{\begin{tabular}[c]{@{}c@{}}P-GPT + MTrans + SA + \\ (Joint Training)\end{tabular}}} & \textbf{68.40} & \textbf{58.70} & \textbf{53.50} & \textbf{47.90} & \textbf{8.1629} \\
& \textit{\textbf{\begin{tabular}[c]{@{}c@{}}P-GPT + Mul-Trans + SA + \\ (Joint Training)\end{tabular}}} & \textbf{69.00} & \textbf{59.10} & \textbf{54.40} & \textbf{49.30} & \textbf{8.5371} \\ \hline
\end{tabular}
\end{adjustbox}
\caption{Results of our proposed approach and the existing baselines }\label{sota}
\end{table}
\subsubsection{Comparison to the Existing Approaches:} In Table \ref{sota}, we present the evaluation results of our proposed framework in comparison to the existing approaches. From the table, it is clearly evident that the use of the slot values improves the generation performance compared to the existing approaches that do not employ slot information for generation. The BLEU-4 score shows an improvement of more than 20 points compared to the unimodal baselines, such as Seq2Seq and HRED networks. 

By using the images, we see that the existing approaches have shown a notable gain in performance as opposed to the unimodal baselines. By using the pre-trained GPT embeddings and slot information, we outperform the best-performing framework \cite{he2020multimodal}, with a BLEU score of 8\%. From this, it can be concluded that slot information assists in correctly responding to user demands and providing interesting and informative responses.

\begin{table}[h]
\centering
\begin{adjustbox}{max width=\linewidth}
\begin{tabular}{c|c|c|c|ccc}
\hline
\multicolumn{2}{c|}{\textbf{Model }} & \textbf{SA} & \textbf{P-GPT} &\textbf{F} & \textbf{R} & \textbf{SC} \\ \hline
\multirow{6}{*}{\textbf{\begin{tabular}[c]{@{}c@{}}Unimodal\\ Baselines\end{tabular}}} & \multirow{3}{*}{\textit{HRED}} & - & - & 3.43 & 3.27 & 2.91 \\
 &  & $\surd$ & - & 3.51 & 3.36 & 3.03 \\ 
 &  & $\surd$ & $\surd$ & 3.57 & 3.43 & 3.12 \\ \cline{2-7} 
 & \multirow{3}{*}{\textit{HRED + Kb}} & - & - & 3.61 & 3.49 & 3.17 \\
 &  & $\surd$ & - & 3.69 & 3.55 & 3.23 \\  
 &  & $\surd$ & $\surd$ & 3.78 & 3.63 & 3.38 \\ \hline
\multirow{3}{*}{\textbf{\begin{tabular}[c]{@{}c@{}}Multimodal\\ Baselines\end{tabular}}} & \multirow{2}{*}{\textit{MHRED}} & - & - & 3.85 & 3.72 & 3.49 \\
 &  & $\surd$ & - & 3.91 & 3.80 & 3.57 \\ 
 &  & $\surd$ & $\surd$ & 3.99 & 3.85 & 3.68 \\ \hline
\multirow{9}{*}{\textbf{\begin{tabular}[c]{@{}c@{}}Proposed\\ Approaches\end{tabular}}} & \multirow{2}{*}{\textit{MHRED + Kb}} & - & - & 4.08 & 3.87 & 3.71 \\
 &  & $\surd$ & - & 4.15 & 3.92 & 3.78 \\
 &  & $\surd$ & $\surd$ & \textbf{4.16} & \textbf{4.02} & \textbf{3.82} \\ \cline{2-7} 
 & \multirow{2}{*}{\textit{MTrans + Kb}} & - & - & 4.29 & 4.02 & 3.87 \\
 &  & $\surd$ & - & 4.36 & 4.18 & 4.05 \\
 &  & $\surd$ & $\surd$ & \textbf{4.42} & \textbf{4.20} & \textbf{4.13} \\ \cline{2-7}
  & \multirow{2}{*}{\textit{Mul-Trans + Kb}} & - & - & 4.33 & 4.14 & 3.96 \\
 &  & $\surd$ & - & 4.42 & 4.25 & 4.12 \\
 &  & $\surd$ & $\surd$ & \textbf{4.53} & \textbf{4.37} & \textbf{4.22} \\\hline
\end{tabular}
\end{adjustbox}
\caption{Evaluation results using human evaluation metrics. Here, SA: Slot Attention, F: Fluency, R: Relevance, SC: Slot Consistency}  
\label{human_res}
\end{table}

The \textit{MHRED} baseline \cite{saha2018towards} merely concatenates the textual and visual information for generating responses which have lower BLEU scores in comparison to the proposed framework. In UMD \cite{cui2019user}, the authors used attention guided hierarchical recurrent encoder-decoder framework for generating responses. Also, enhanced visual representation achieved with the help of a taxonomy attribute tree was used for correct response generation. 

It is visible that by explicitly using the slot information in the \textit{MHRED} network, it outperforms the UMD framework, giving a boost of more than 20\% in BLEU-4 scores. The improvement is mainly due to the usage of pre-trained DialoGPT embeddings and slot attention that provide enhanced contextual information compared to the recurrent encoders. The \textit{MTrans} framework yields superior performance, proving the efficacy of Transformers as opposed to the recurrent networks. 

The OAM \cite{chauhan2019ordinal} network focuses upon the attributes and position of the images, and employs the MFB fusion technique to obtain the non-linear interaction between the modalities for generating coherent responses with a NIST score of 4.3236. Our proposed transformer-based approach attains around 4\% gain in the NIST score in contrast to the OAM framework. Though MATE \cite{he2020multimodal} exploits the transformer as encoder-decoder, our proposed approach still performs well in comparison. This is primarily because of the slot attention mechanism that correctly focuses on the correct attributes of the product, and makes the responses coherent, informative and interactive.



\begin{table}[h]
\centering
\begin{adjustbox}{max width=\linewidth}
\begin{tabular}{c|c|c}
\hline
\textbf{Models} & \textbf{SIMCC-Furniture} & \textbf{SIMCC-Fashion} \\ \hline
\textit{LSTM} & 0.022 & 0.022 \\
\textit{HAE} & 0.075 & 0.059 \\
\textit{HRE} & 0.075 & 0.079 \\
\textit{MN} & 0.084 & 0.065 \\
\textit{T-HAE} & 0.044 & 0.051 \\ \hline
\textit{\begin{tabular}[c]{@{}c@{}}Mul-Trans\\ (our)\end{tabular}} & 0.086 & 0.080 \\ \hline
\end{tabular}
\end{adjustbox}
\caption{Results of BLEU score on SIMCC dataset}  
\label{simcc}
\end{table}

In Table \ref{simcc}, we present the evaluation results of different frameworks on the SIMCC dataset \cite{moon2020situated} along with our proposed model. As shown in the table, our proposed framework performs slightly better than the best performing \textit{MN} (memory network) for SIMCC-Furniture data and \textit{HRE} (Hierarchical Recurrent Encoder) for SIMCC-Fashion data, respectively. One of the main reasons is that we use slot-based attention that helps in focusing on the attributes and the transformer framework which is more robust than the RNN framework. The effectiveness of our proposed framework is evidenced through another multimodal dataset, \textit{viz.} SIMCC, which ensures that it can be used for similar other datasets. 

\subsubsection{Results of Human evaluation:}
Along with the automatic evaluation, we also report the results of the manual evaluation in Table \ref{human_res}. The results of the manual evaluation are in consonance with the automatic evaluation results. The fluency of the proposed framework is highest in comparison to all the baselines. By providing pre-trained embedding information, slot attention, and the external knowledge base, the responses are complete, thereby being grammatically fluent. 

The proposed framework's relevance score is maximum, ensuring that the responses are coherent with the given dialogue context. As can be seen from the table, the relevance scores of the multimodal frameworks are higher than the unimodal networks as image information helps provide the full context for the generation of coherent responses. Also, the inclusion of the knowledge base improves the score in all the baselines and the proposed network. As our current work's primary objective is to generate more informative responses in accordance with the extracted slot information, we see that the scores of the slot consistency metric increase with the incorporation of slot attention. 

Also, the models having the knowledge base information outperform the other frameworks mainly because the external knowledge constitutes the important slot information for a particular dialogue, hence boosting the generation of responses. Utilizing transformers instead of GRU has been shown to improve the quality of responses as is evident from the results depicted in Table \ref{human_res}. The fluency of the responses takes a jump from \textit{4.16} score to \textit{4.42} proving the efficacy of Transformers to generate better responses. Similarly, the responses are relevant to the conversational history making the responses consistent with the ongoing dialogue. 

As the primary goal of our work is to make the responses interactive by using the correct slot information, hence on evaluating we see that compared to the \textit{MHRED} network, \textit{Mtrans} shows a performance gain in the slot consistency metric as well. \textit{Mul-Trans} model in consonance with the automatic evaluation results shows better performance compared to all the baselines in terms of manual evaluation as well. Visual representation from transformers has achieved providing better responses that are fluent and contextually coherent as well.  

Previously, the author in \cite{su2019dual} employed a dual learning mechanism to jointly address the NLU and NLG tasks in a dialogue system. Similarly in \cite{tseng2020generative}, the author investigates a generative RNN framework for both tasks. Direct comparison to the existing frameworks has not been shown in our current work, firstly because these networks are trained solely on textual data. Secondly, in our work, we focus on extracting the slot information from the user utterance to facilitate the network in enhancing the next textual system response generation. In contrast, in the existing frameworks, NLU and NLG are performed on a single utterance only (i.e. the extracted slot is used to generate the same utterance while we generate the next response in the dialogue). Therefore, our work is novel as we exploit multimodal sources for both NLU and NLG tasks in a user-system dialogue setting. 
\begin{figure}[htbp]
    \centering
    \begin{adjustbox}{max width=\linewidth}
    \includegraphics{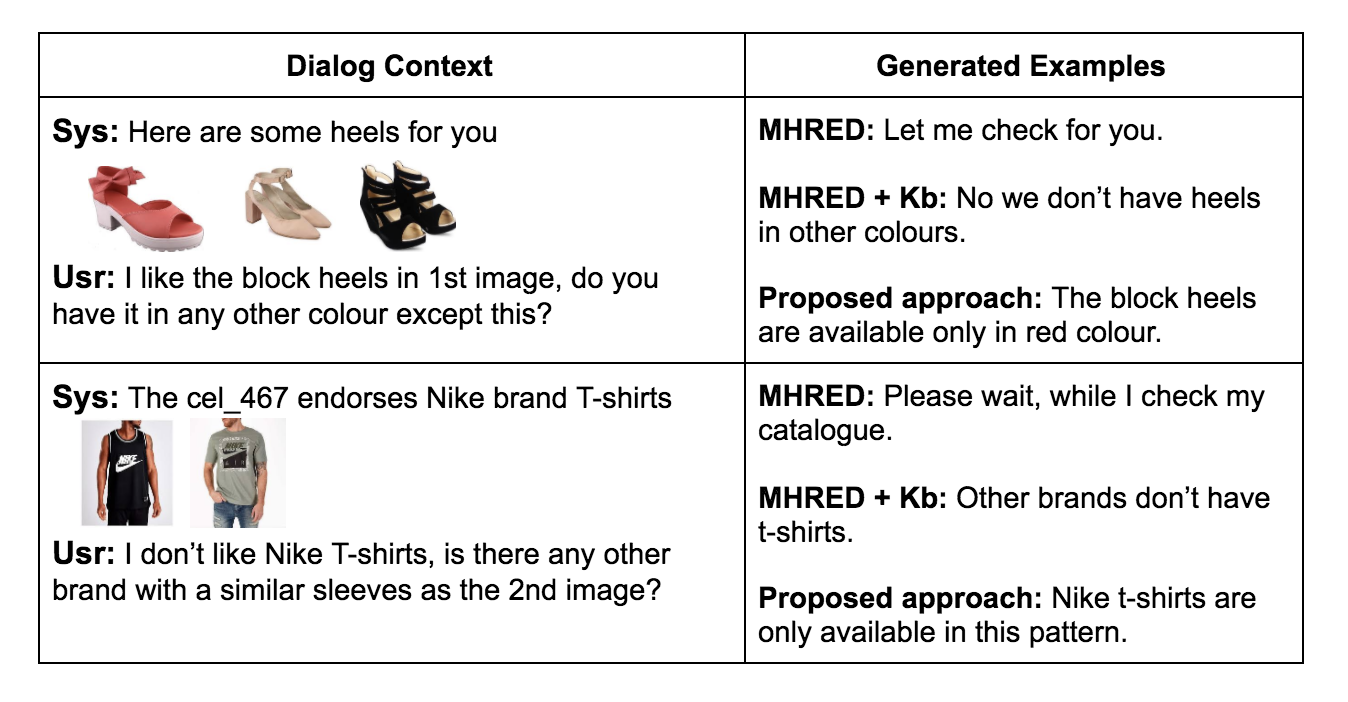}
    \end{adjustbox}
    \caption{Examples of generated responses by baseline and proposed framework}
    \label{gen_eg}
\end{figure}
\subsection{Case Studies and Error Analysis}
In Figure \ref{gen_eg}, we provide a few examples generated by our proposed framework and the baselines. In Example 1 from the Figure, the proposed approach generates a more informative and relevant response by including the correct slot values such as \textit{block heels} as desired by the user. Similarly, in the second example, the ability to generate the correct brand \textit{Nike} and type \textit{t-shirts} of the product makes the response diverse and interactive as opposed to the responses generated by the baseline networks. 

The baseline networks without the slot information tend to generate safer responses that lack specific patterns, and brand information in them. While the examples generated by the proposed approach it is obvious that the correct slot information assists in generating better responses and providing specified and desired products to the user thereby increasing customer satisfaction and increasing customer retention.

We closely analyze the outputs of the generated response to be aware of the errors made by the proposed dialogue generation framework. The common errors made by the model are: 
\begin{itemize}
    \item \textbf{Erroneous image selection:} The model is sometimes incompetent in selecting the images having contextual information of more than 5 turns, thereby generating incorrect responses in some cases. There are also cases, where due to the discussion of multiple images in the conversational history, wrong images get selected, making the responses incorrect.
    \item \textbf{Additional Information:} The model sometimes generates additional/extra information in the case of attributes for a few products. For example, \textbf{Gold:} \textit{The material of the trousers is cotton.}, \textbf{Predicted:} \textit{The trousers have cotton polyester material with check patterns.} This is mainly due to the fact that the conversation history has this additional information which also gets incorporated into the responses.
    \item \textbf{Repetition:} Sometimes the baseline and proposed frameworks generate words that are repeated throughout the response. Also, unknown words due to fewer instances in the training data get generated as <unk> tokens in the responses. For example, \textbf{Gold:} \textit{The 3rd image belongs to the Fossil brand with a blue dial.}, \textbf{Predicted:} \textit{The 3rd image has <unk> <unk> <unk>.} 
    \item \textbf{Slot mismatch:} Sometimes due to multiple slot information in the ground truth response and the conversational history, the proposed framework at times gets confused and generates responses that have incorrect slot information. For example, \textbf{Gold:} \textit{The red synthetic material top with bell sleeves will look good with the trousers.}, \textbf{Predicted:} \textit{The trousers have red material with bell patterns.}
\end{itemize}

The above-mentioned errors could be minimized by including better image encoders to capture visual representations. In addition, fusion techniques that could capture non-linear interactions between the modalities could help with the errors of type \textit{slot mismatch}. For \textit{repetition} errors we plan to investigate better pre-trained language models that could also handle the additional information issue. 

\section{Conclusions and Future Work}
With the progress in artificial intelligence, dialogue systems have reached new paradigms. Narrowing the gap between vision and language, multimodal conversational systems have gained immense popularity. Complementary information in the form of images, audio, or videos to the unimodal (text) systems has helped build robust systems. Task-oriented dialogue systems focus on assisting humans by assisting them to achieve their desired goals. Response generation is a crucial component in every dialogue system. 

Our current work emphasizes the task of generating responses in a multimodal dialogue system. In this paper, we have proposed an end-to-end framework capable of extracting slot values from user utterances and generating a suitable response. For improving the performance of slot extraction, we apply a self-attention mechanism on the utterances so that the appropriate slot values get focused. In addition to this, we also employ self-attention on Knowledge Base (KB) and use this attended representation to assist in generation. Furthermore, we pre-train DialoGPT Language Model onto a Multi-modal Dialog dataset. 

This effectively learns the previous dialogue context along with the current user utterance and system response. Contextual embeddings trained using DialoGPT are passed as the input to the model. We evaluated our proposed approach on the Multi-modal Dialog dataset and have shown significant performance improvement. Our proposed approach focused on essential slot information in qualitative and quantitative metrics, thereby improving the generation.

Our current approaches focus most on the text and little on the image. In the future, we wish to enhance the performance by proposing methodologies that utilize information from the images. These methods include fusion techniques capable of incorporating information from both image and text, thereby improving Multimodal dialogue systems' performance.

\bibliographystyle{splncs04}
\bibliography{paper}
\end{document}